\documentclass[lettersize,journal]{IEEEtran}
\usepackage{algorithmic}
\usepackage{algorithm}
\usepackage{array}
\usepackage[caption=false,font=normalsize,labelfont=sf,textfont=sf]{subfig}
\usepackage{textcomp}
\usepackage{stfloats}
\usepackage{url}
\usepackage{verbatim}
\usepackage{graphicx}
\usepackage{cite}
\usepackage{multirow}
\usepackage{makecell}
\usepackage{xcolor}
\usepackage{amsmath}
\usepackage{amsfonts}
\usepackage{pifont}
\newcommand{\cnum}[1]{\ding{\numexpr191+#1\relax}}
\usepackage{booktabs}
\newif\ifshownotes
\shownotestrue      % <-- show notes
\newcommand\etal{\emph{et al.}}

\IEEEoverridecommandlockouts

\usepackage[font=small]{caption}
\begin{document}
%\bstctlcite{MyBSTcontrol}

\title{MS-MEM: Multi-Skill Manipulation-Enhanced Mapping\\
via Uncertainty- and Disturbance-Aware Action Selection}

\author{
Yitian Shi\textsuperscript{1,*},
Jesper Mücke\textsuperscript{2,*},
Nils Dengler\textsuperscript{3},
Sicong Pan\textsuperscript{2},
Rania Rayyes\textsuperscript{1},
and Maren Bennewitz\textsuperscript{2}
\thanks{\textsuperscript{1}Karlsruhe Institute of Technology, Germany.}
\thanks{\textsuperscript{2}University of Bonn, Germany.}
\thanks{\textsuperscript{3}Technical University of Darmstadt, Germany.}
\thanks{This work was supported by the German Federal Ministry of Research,
Technology and Space (BMFTR) under the Robotics Institute Germany (RIG),
the DFG through project SFB~1574 (Project No.~471687386), and the Ministry
of Science, Research and the Arts of the State of Baden-Württemberg through
the InnovationCampus Future Mobility.}
\thanks{S. Pan and M. Bennewitz are also affiliated with the Lamarr Institute
for Machine Learning and Artificial Intelligence and the Center for Robotics,
Bonn, Germany.}
\thanks{*Equal contributionbu.}
}
\markboth{}%
{Shell \MakeLowercase{\textit{et al.}}: A Sample Article Using IEEEtran.cls for IEEE Journals}
\maketitle
\begin{abstract}
Accurate scene understanding in confined, cluttered spaces such as shelves is essential for service robots, as many everyday tasks require them to locate and retrieve objects reliably. Yet, it remains challenging due to severe occlusions, restricted accessibility, and the need to avoid excessive scene changes.
In this paper, we propose Multi-Skill Manipulation-Enhanced Mapping~(MS-MEM), an evidential framework for uncertainty-aware mapping that integrates active viewpoint selection, object pushing, and grasping.
\mbox{MS-MEM} combines scene-level metric-semantic evidential belief estimators with an uncertainty-aware grasp representation. 
This representation is learned using a novel full-evidential grasp estimator that models both grasp affordance and orientation uncertainty.
In our framework, candidate perception and manipulation actions are evaluated within a unified action selection pipeline using a common information gain criterion. 
For manipulation actions, we further introduce a collateral disturbance constraint (CDC) that discourages excessive changes to confident regions of the scene belief. This enables MS-MEM to select actions that effectively reduce map uncertainty while limiting collateral scene changes.
Experimental results show that, compared with single-skill and unconstrained baselines that ignore scene disturbance, \mbox{MS-MEM} achieves higher mapping accuracy while substantially reducing scene disturbance, highlighting the synergistic effects of active viewpoint selection, push, and grasp actions.
\end{abstract}


\section{Introduction}
Robots operating in cluttered and confined environments often need to acquire task-relevant scene information under severe occlusion. This information may range from target localization and accessibility for object retrieval~\mbox{\cite{efendi2025technological, shi2025viso, li2016act}} to semantic-aware mapping for automated inventory~\mbox{\cite{marques2025map, huang2021mechanical}}.
In these scenarios, heavy occlusions and restricted access often limit the informativeness of visual observations, making passive or active sensing alone insufficient for reliable scene interpretation~\cite{bohg2017interactive}.
Perception and manipulation need, therefore, to be tightly coupled, as successful manipulation depends on accurate estimates of the environment, while accurate perception may require targeted interaction with the environment to reveal occluded regions and reduce uncertainty.

To address these challenges, early work assumes a fixed viewpoint during manipulation~\cite{danielczuk2019mechanical, huang2021mechanical}, limiting observability under occlusion and preventing active view selection. 
Other work \cite{shi2025viso} combines active vision with grasping to retrieve hidden objects in unrestricted cluttered scenes.
More recently, Marques~\etal~\cite{marques2025map} proposed  Manipulation-Enhanced Mapping~(MEM) that formulates mapping as a sequential decision-making process under uncertainty, alternating between next-best view selection and non-prehensile interaction while updating a learned metric-semantic belief map.
However, MEM is restricted to pushing as its sole interaction primitive. 
This can be suboptimal, as pushing may unnecessarily disturb object configurations, invalidate parts of the accumulated map, and miss opportunities where more selective actions, such as grasping, would yield more efficient and less disruptive information gain.
Moreover, recent work on synergistic push–grasp manipulation~\cite{zeng2018learning, hu2025push, xu2021efficient} in unconfined areas shows that pushing and grasping naturally complement each other in cluttered environments.
In particular, pushing can separate objects, create free space, and reveal occluded regions, whereas grasping can selectively remove reachable occluders at the visibility frontier while limiting collateral disturbance to neighboring
objects.
In the context of MEM, this complementarity can improve mapping performance by exposing occluded regions while enabling more selective manipulation of the scene.

\begin{figure}[]
    \centering
    \hspace*{-0.03\linewidth}
    \includegraphics[width=1.0\linewidth, trim= 0 0 0 5, clip]{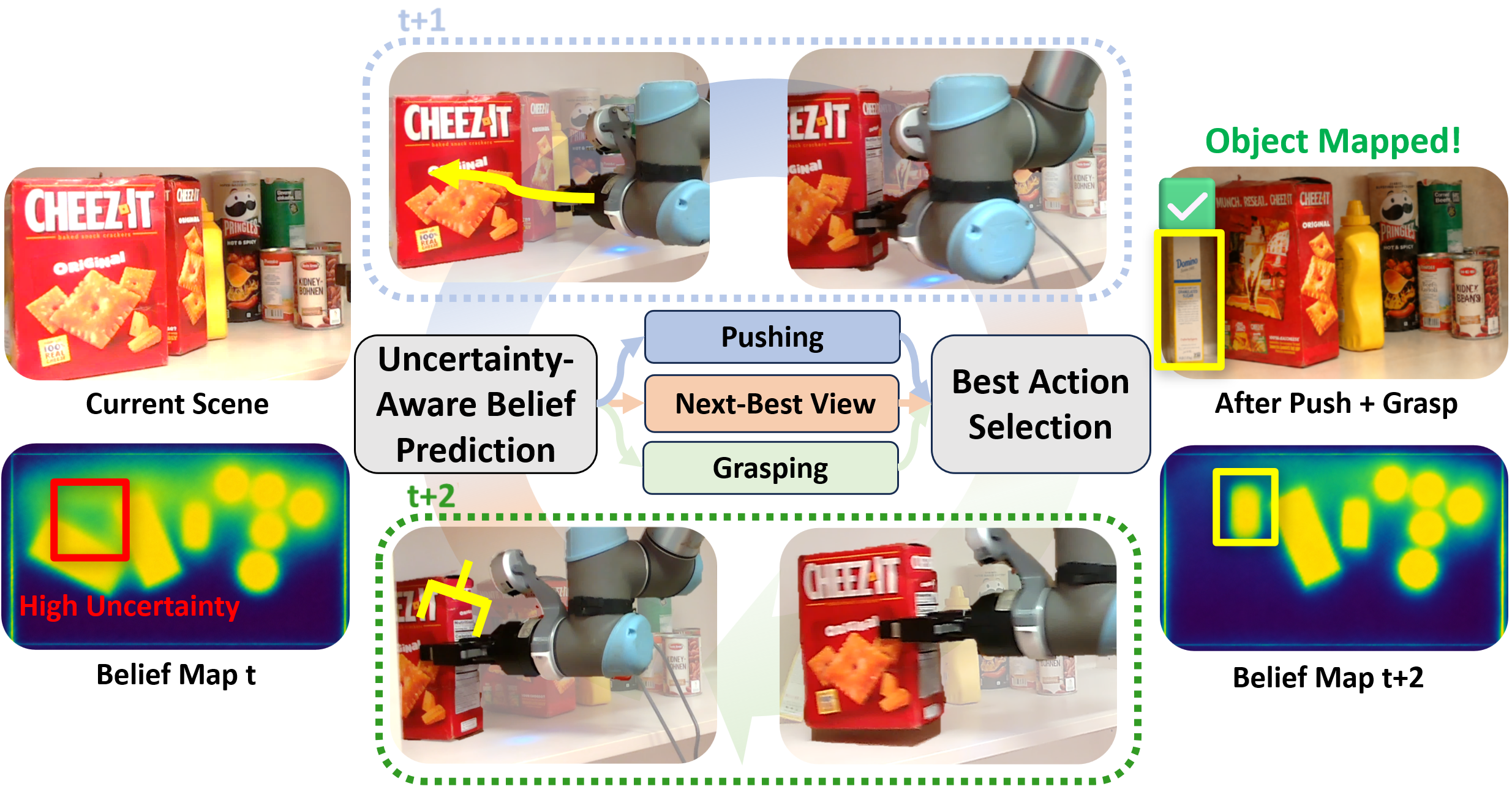}
   \caption{Overview of MS-MEM: Uncertainty-aware grasping, pushing, and active vision jointly remove the occluder (Cheez-It box) in a confined shelf so that a hidden object (yellow box) is identified and forwarded to the uncertainty-aware mapping.
   }
   \vspace{-2mm}
    \label{fig:teaser}
\end{figure}

Therefore, we propose Multi-Skill Manipulation-Enhanced Mapping~(MS-MEM), a framework for active metric-semantic mapping in cluttered and confined environments. As illustrated in Fig.~\ref{fig:teaser}, \mbox{MS-MEM} autonomously selects among active view planning, pushing, and grasping according to their expected contribution to uncertainty reduction and scene preservation. To support grasp planning under partial observability, we introduce Full-Evidential vMF-Contact~(FE-vMF), which extends vMF-Contact~\cite{shi2025vmf} to a comprehensive evidential grasp representation that jointly models grasp affordance and orientation uncertainty. 
This facilitates uncertainty-guided multi-view grasp fusion, enabling consistent grasp refinement as the map belief evolves across diverse viewpoints. All candidate actions across skills are evaluated under a common occlusion-aware information gain objective, incorporating a collateral disturbance constraint~(CDC) to explicitly penalize unnecessary scene changes.
This enables the selection of highly informative actions while preserving the reliability of previously accumulated map evidence.
Our experimental results demonstrate that combining pushing and grasping yields maps that are more geometrically complete and semantically accurate than single-skill alternatives.
Furthermore, incorporating the CDC penalty maintains competitive mapping performance while substantially reducing unnecessary object position changes induced by manipulation.

%Our experiments demonstrate the complementary benefits between pushing and grasping within MS-MEM and show that jointly reasoning over pushing and grasping yields more complete and semantically consistent maps than corresponding single-skill comparisons. Moreover, when constraining scene alteration during action selection, our method achieves a significantly lower position change while better mapping performance than MEM, showing that the proposed disturbance-aware objective improves scene preservation by discouraging overly disruptive manipulations, especially pushes.
% In summary, our contributions are the following: (i) We propose Multi-Skill Manipulation-Enhanced Mapping~(MS-MEM), which extends MEM \cite{marques2025map} by joint reasoning over active view selection, pushing, and grasping actions for uncertainty-aware occlusion reduction with scene disturbance as the additional constraint; (ii) We introduce a hierarchical framework that systematically incorporates a novel evidential grasp representation learning module into the metric-semantic belief representation of MEM; (iii) We develop an uncertainty-calibrated grasp selection and fusion mechanism to aggregate grasp hypotheses over time for consistent action selection; (iv) We propose a unified decision-making objective across skills, namely the Disturbance- and Occlusion-aware Information Gain (DOIG), which maximizes the information gain for mapping while constraining scene disturbance. 
In summary, our main contributions are as follows:
(i) We propose MS-MEM, a multi-skill manipulation-enhanced mapping
%adaptation of the MEM framework~\cite{marques2025map} 
for uncertainty-aware mapping in cluttered and confined scenes, enabling uncertainty-informed action selection over active viewpoint changes, pushing, and grasping. (ii) We develop a comprehensive evidential grasp learning framework that jointly models uncertainty in grasp affordance and orientation and employs uncertainty-guided multi-view fusion to ensure consistent grasp selection over time. (iii) We design a unified multi-action decision objective based on Disturbance- and Occlusion-aware Information Gain~($\mathrm{DOIG}$), enabling direct comparison across perception and manipulation actions while penalizing unnecessary scene changes.

\section{Related Work}

\subsection{Manipulation for Mechanical Search and Mapping}
To explore, map, and retrieve objects in cluttered environments, robots must actively manipulate the scene to reveal occluded or unknown regions~\cite{bohg2017interactive}. 
Mechanical search addresses this as a sequential decision-making problem, studying long-horizon interactions for localizing and retrieving hidden objects under partial observability~\cite{li2016act, xiao2019online}. 
Early work studied multi-step target retrieval in tabletop clutter using pushing and grasping primitives~\cite{danielczuk2019mechanical, xiao2019online}, while later approaches considered confined shelf settings, estimating target occupancy and planning rearrangements to reveal hidden objects~\cite{huang2021mechanical}. 
More recently, semantic mechanical search~\mbox{\cite{sharma2023semantic, shi2025viso}} further incorporates language and vision priors to infer likely target locations from contextual cues.

In comparison, recent work extends these ideas from target retrieval to manipulation for accurate scene mapping. 
For this, \cite{dengler2023viewpoint} combined viewpoint planning with pushing to uncover occluded regions in confined shelves. 
Building on top, Manipulation-Enhanced Mapping~(MEM) formulates mapping as a Partially Observable Markov Decision Process (POMDP) over uncertainty-aware metric-semantic belief maps~\cite{marques2025map, dengler2025efficient}, using uncertainty for action selection through continuous Next-Best View (NBV) planning and for push sampling to target occlusion-critical objects while reducing unnecessary disturbance \cite{liu2024efficient}. Despite these advances, existing mapping methods remain limited to pushing as the only manipulation primitive.

\subsection{Push-Grasp Synergy}
The synergistic effect between push and grasp actions has been widely studied as a way to improve manipulation in cluttered environments~\cite{efendi2025technological, zeng2018learning, xu2021efficient, hu2025push}. 
While pushing can rearrange clutter\cite{bejjani2021learning} or separate objects~\cite{hermans2012guided}, grasping enables targeted object removal once suitable opportunities are available~\cite{newbury2023deep}. 
Rather than treating these primitives independently, prior work has shown that jointly reasoning about pushing and grasping improves manipulation efficiency and robustness, particularly when objects are densely packed, partially occluded, or initially difficult to grasp
%In this context, several approaches use pushing to facilitate subsequent grasping by exposing object boundaries, increasing free space, or changing object poses to improve grasp feasibility
~\cite{zeng2018learning, xu2021efficient}. 
Others learn policies that decide between pushing and grasping based on the expected manipulation outcome, allowing the robot to exploit pushing when grasp success is unlikely and grasping when direct removal is feasible~\cite{hu2025push}. 
This line of work demonstrates that the complementary roles of pushing and grasping can reduce failed interactions and improve task completion in clutter. 

In contrast to goal-oriented manipulation, our work leverages push-grasp synergy for uncertainty reduction in active mapping, where actions are selected not only for manipulation success but also for their expected contribution to scene understanding and minimizing scene changes.

\subsection{Uncertainty-Aware Grasp Learning}
%To execute reliable grasps in cluttered and partially observed scenes, robots must reason about not only the grasp quality, but also the uncertainty in the predicted grasp.
Learning-based 6-DoF grasp synthesis \cite{newbury2023deep} has been driven by large-scale datasets and contact-centric grasp representations, which enable reliable grasp prediction in cluttered scenes~\cite{sundermeyer2021contact, shi2025vmf}. 
Recent work further models grasp distributions explicitly to improve robustness in clutter and bin-picking~\cite{liu2024efficient}. 
Along this line, uncertainty-driven online grasp learning uses predictive uncertainty to guide exploration under distribution shift~\cite{shi2024uncertainty}, while vMF-Contact~\cite{shi2025vmf} proposes an evidential formulation based on von Mises--Fisher (vMF) distributions to represent directional uncertainty in probabilistic contact-grasp synthesis. 
Multi-view uncertainty has also been explored for active learning~\cite{gilles2025metamvuc}, while remaining limited to a small number of predefined viewpoints in table-top settings.

In contrast, we incorporate a full-evidential grasp representation into our framework, enabling grasp actions to be assessed not only by executability and uncertainty, but also by their expected contribution to reducing evidential map uncertainty. Moreover, by building on MEM, our method naturally supports multi-view reasoning during sequential interaction, allowing grasp decisions to benefit from observations accumulated across changing viewpoints.

\section{Preliminary}

\subsection{Manipulation-Enhanced Mapping (MEM)}
\label{sec:mem}
To efficiently reconstruct cluttered and partially observable scenes, we build our approach on Manipulation-Enhanced Mapping (MEM)~\cite{marques2025map, dengler2025efficient}, which formulates mapping as a sequential decision-making process under uncertainty.
At time $t$, the scene is represented by a metric-semantic belief \mbox{$\Phi_t = \{\Phi_t^{O}, \Phi_t^{S}\}$}, where $\Phi_t^{O}$ denotes a volumetric occupancy belief and $\Phi_t^{S}$ a 3D semantic belief over known classes.

Since occlusions and partial observations make both occupancy and semantic estimates ambiguous, explicitly representing uncertainty is essential for selecting informative sensing and manipulation actions. 
MEM therefore adopts an evidential representation $\lambda_t = \{\lambda_t^{O}, \lambda_t^{S}\}$, where
$\lambda_t$ stores Beta and Dirichlet parameters for each map element, respectively.
These parameters define the posterior distributions over the metric-semantic
belief $\Phi_t=\{\Phi_t^{O},\Phi_t^{S}\}$, whose expected values are used as the occupancy and semantic belief maps.

In particular, the occupancy probability of each voxel $u$ is modeled by a Beta distribution with parameters $\alpha_{t,u}^{O}, \beta_{t,u}^{O}$:
\begin{equation}
\Phi_t^{O}[u] \sim \mathrm{Beta}(\alpha_{t,u}^{O}, \beta_{t,u}^{O}), \quad
\mathbb{E}[\Phi_t^{O}[u]] = \frac{\alpha_{t,u}^{O}}{\alpha_{t,u}^{O} + \beta_{t,u}^{O}},
\label{eq:occ_beta}
\end{equation}
the semantic probabilities are modeled by Dirichlet distributions with concentrations $[\eta_{t,u,c}^{S}]_{c=1}^{N_{\mathrm{cls}}}$ under $N_{\mathrm{cls}}$ classes:
\begin{equation}
\Phi_t^{S}[u] \sim \mathrm{Dir}([\eta_{t,u,c}^{S}]_{c=1}^{N_{\mathrm{cls}}}), \quad
\mathbb{E}[\Phi_{t,c}^{S}[u]] = \frac{\eta_{t,u,c}^{S}}{\sum_{c'} \eta_{t,u,c'}^{S}}
\end{equation}
This evidential formulation \cite{gao2025comprehensive} enables calibrated reasoning about uncertainty and occlusions in the scene.
To update the belief state, we approximate the new state at $t+1$ using learned Calibrated-Neural Accelerated Belief Updates~(CNABU) networks~\cite{marques2025map}, i.e., given an action $a_t$ and observation $o_t$, the evidential state is updated as
\begin{equation}
\lambda_{t+1} \leftarrow \sigma_o(\lambda_t, o_t, a_t), \quad
\lambda_{t+1} \leftarrow \sigma_p(\lambda_t, a_t),
\label{eq:cnabu}
\end{equation}
where $\sigma_o$ and $\sigma_p$ denote observation and push-specific networks to update the beliefs, respectively. 
This enables efficient prediction of post-action beliefs, which is later used to evaluate candidate actions via expected information gain.

\subsection{Hierarchical Bayesian Grasp Representation}
To reason about grasp feasibility and uncertainty under partial observability, we formulate 6-DoF grasp synthesis as a hierarchical Bayesian inference problem following~\mbox{\cite{liu2024efficient, shi2025vmf}}.
Concretely, we adopt the contact-based evidential grasp representation of 
%Shi~\etal~
\cite{shi2025vmf}, which models 6-DoF grasps at a contact point $\boldsymbol{c}$ in a hierarchical manner:
 \begin{equation}
     p(\boldsymbol{g}|\boldsymbol{c}) =
     \int \underbrace{p(q | \boldsymbol{c})}_{\text{Affordance}} \cdot
     \underbrace{p(\boldsymbol{b} | q, \boldsymbol{c})}_{\text{Baseline}} \cdot
     \underbrace{p(\boldsymbol{a} | \boldsymbol{b}, q, \boldsymbol{c})}_{\text{Approach}} 
     % \cdot \underbrace{p(w | \boldsymbol{b}, q, \boldsymbol{c})}_{\text{Width}} \, 
%     dq
%     \label{eq:bayes_grasp}
\end{equation}
Here, following the contact-based grasp representation of~\cite{sundermeyer2021contact}, a grasp $\boldsymbol{g}$ is decomposed into an affordance score $q$, indicating the graspability of the contact point, and two orthogonal orientation components: the approach direction $\boldsymbol a$, describing the direction from which the gripper approaches the object, and the baseline direction $\boldsymbol b$, representing the gripper closing direction between the two fingers. The directional uncertainty for the baseline $\boldsymbol b$ is modeled by~\mbox{von Mises-Fisher~(vMF)} distributions:
\begin{equation}
p(\boldsymbol{b} \mid \boldsymbol{\mu}, \kappa) = \mathcal{C}(\kappa) \exp(\kappa \boldsymbol{\mu}^\top \boldsymbol{b}),
\end{equation}
where ${\boldsymbol{\mu}} \in S^2$ is the mean direction and $\kappa > 0$ is the concentration parameter, normalized by $\mathcal{C}(\kappa) = \frac{\kappa}{4\pi \sinh \kappa}$.
This enables uncertainty-aware grasp prediction and supports multi-view aggregation through Bayesian updates~\cite{shi2025viso}. 

\section{Methodology}

\begin{figure*}[]
\centering
    \includegraphics[width=\linewidth]{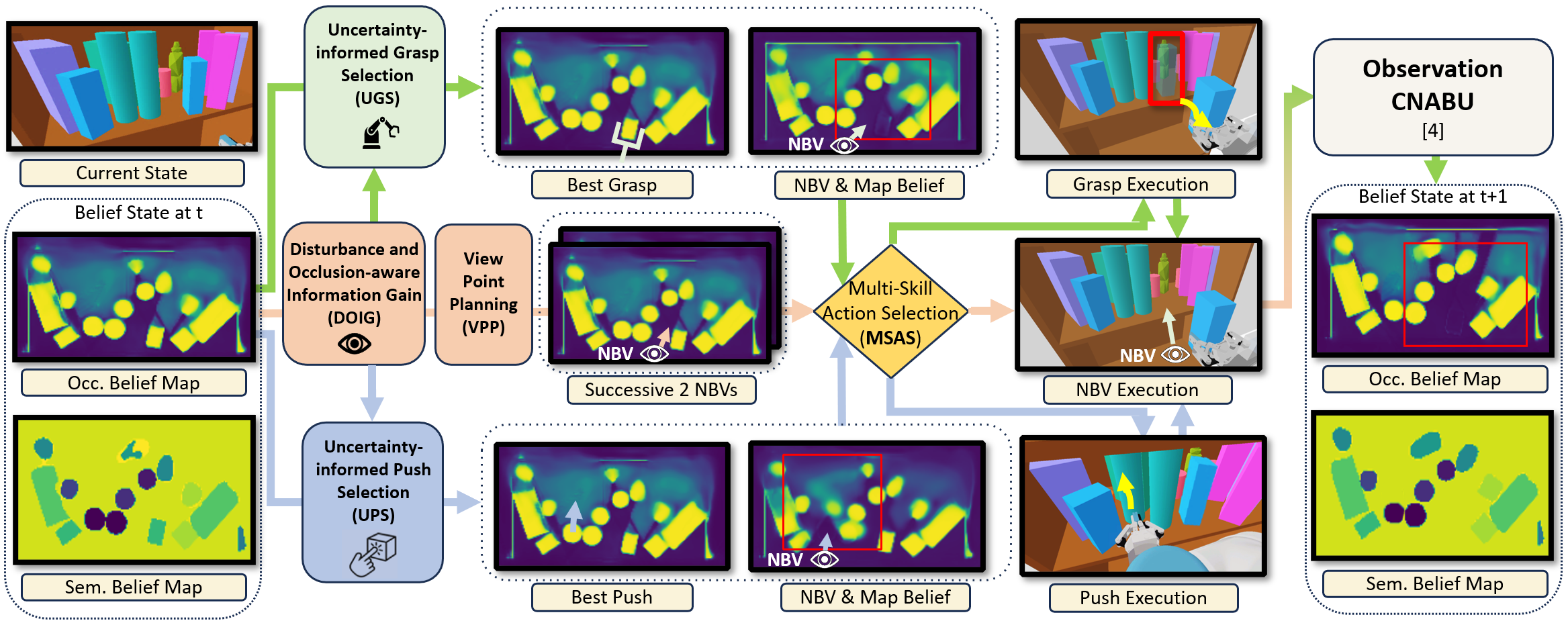}
    \caption{Given the current evidential belief state $\Phi_t$, our MS-MEM~framework evaluates three action modalities in parallel: Uncertainty-Informed Grasp Selection (\textbf{UGS}), active viewpoint selection through Next-Best View~(\textbf{NBV}) planning, and Uncertainty-Informed Push Selection (\textbf{UPS}). The Disturbance- and Occlusion-aware Information Gain (\textbf{DOIG}) objective evaluates successive NBV candidates~\cite{marques2025map} and supports both UGS and UPS by assessing the information gain and collateral disturbance associated with their predicted post-manipulation beliefs. The Multi-Skill Action Selection (\textbf{MSAS}) module then compares the best grasp, push, and viewpoint candidates under the shared DOIG objective and selects the action $a_t^*$ to execute. When a manipulation action is selected, its associated NBV is subsequently executed, and the resulting observation is integrated by the observation CNABU $\sigma_o$ to update the belief state.} %\todo{add Map Belief to 2 NBVs}}
    \label{fig:main_pipeline}
    \vspace{-2mm}
\end{figure*}

We now present MS-MEM, a unified multi-skill framework for uncertainty-aware mapping through joint reasoning over perception and manipulation, which integrates grasping, non-prehensile pushing, and active view selection into a unified decision-making process. 
As illustrated in Fig.~\ref{fig:main_pipeline}, the system operates as a closed-loop framework over a shared evidential scene belief $\Phi_t$. 
%We begin with a high-level overview of the system, and subsequently describe its individual components in detail.
\subsection{Overview}
\label{sec:system_overview}
At each time step, candidate actions from the three modalities, i.e., occlusion-aware viewpoint selection, as well as uncertainty-informed push and grasp selection, are generated and evaluated jointly.
To make all actions comparable, we evaluate their post-action beliefs under a common Disturbance- and Occlusion-aware Information Gain~(DOIG) objective (Sec.~\ref{sec:doig}), which enables direct comparison across sensing and manipulation.

To predict suitable grasp candidates $\boldsymbol{g}_t$, we use the belief~$\Phi_t$ to infer uncertainty-aware grasp hypotheses with our proposed full-evidential vMF-Contact (Sec.~\ref{sec:FE-vMF}) method.
These hypotheses are then further accumulated over time by our novel full-evidential uncertainty-guided multi-view grasp fusion (Sec.~\ref{sec:FE-UMGF}) and evaluated using the DOIG objective.
For the other two considered action types, we generate push actions $\boldsymbol{p}_t$ following~\cite{dengler2025efficient} to target occluded regions based on the uncertainty estimates, while viewpoint candidates~$\boldsymbol{v}_t$ from a set of possible viewpoints~$\mathcal{V}$ are evaluated through classical Next-Best View (NBV) planning~\cite{marques2025map}. 
Finally, after the execution of the chosen actions, the individual CNABU $\sigma_{(\cdot)}$ for each action type updates the evidential map according to the newly acquired observation $o_{t+1}$ or the predicted physical scene change in case of a manipulation.

\subsection{Disturbance- and Occlusion-Aware Information Gain}

\label{sec:doig}
Our novel Disturbance- and Occlusion-aware Information Gain ($\mathrm{DOIG}$) formulation is a shared decision metric that extends volumetric Information Gain~($\mathrm{IG}$)~\cite{delmerico2018comparison} and is used to evaluate and compare all individual action modalities under a common uncertainty-aware objective.
For manipulation actions~\mbox{$a_t \in \{p_t,g_t\}$}, our system first predicts the post-action belief $\tilde{\Phi}_{t+1}^{a_t}$ and then evaluates the expected $\mathrm{IG}$ of the best subsequent viewpoint $v_{t+1}\in\mathcal{V}$. 
For pure viewpoint actions~$a_t\in\mathcal{V}$, the utility combines the immediate information gain with the best subsequent gain after incorporating the resulting observation via NBV planning and observational belief update by~$\sigma_o$ (Eq.~\eqref{eq:cnabu}). 
%In this way, all branches are evaluated over an equivalent two-step horizon, enabling a fair comparison. 

Following~\cite{marques2025map}, we first express the consequences that taking action $a_t$ has to the belief space using Occlusion-aware Information Gain ($\mathrm{OIG}$):
\begin{equation}
\begin{aligned}
\mathrm{OIG}(a_t)
&:= \max_{v_{t+1}\in \mathcal{V}}
\mathrm{IG}(v_{t+1}\mid \tilde{\Phi}_{t+1}^{a_t}) \\
&\quad +
\begin{cases}
\zeta_o \Delta \mathrm{H}(\Phi_t, \tilde{\Phi}_{t+1}^{a_t}),
& a_t \in \{p_t,g_t\} \\[0.3em]
\mathrm{IG}(a_t\mid \Phi_t)
& a_t \in \mathcal{V}
\end{cases}
\end{aligned}
\label{eq:oig}
\end{equation}
For manipulation actions, the semantic entropy difference $\Delta \mathrm{H}(\Phi_t,\tilde{\Phi}_{t+1}^{a_t})$, weighted by~$\zeta_o$, penalizes uncertainty introduced by the predicted scene change: 
\begin{equation}
\Delta \mathrm{H}(\Phi_t,\tilde{\Phi}_{t+1}^{a_t})
:=
\mathrm{H}(\Phi_t)
-
\mathrm{H}(\tilde{\Phi}_{t+1}^{a_t})
\end{equation}
This regularization presents semantic uncertainty caused by object rearrangements, which volumetric $\mathrm{IG}$ does not capture.
%For viewpoint actions, OIG is equivalent to cumulative volumetric IG over two consecutive NBV steps. 
%This asymmetry reflects that viewpoint actions only reduce uncertainty through incremental observation, whereas manipulation actions also alter the scene and may introduce novel uncertainty through their predicted outcomes. 

However, optimizing $\Delta \mathrm{H}(\Phi_t,\tilde{\Phi}_{t+1}^{a_t})$
alone does not necessarily prevent unnecessary changes in regions that are
already mapped with high confidence. 
Moreover, since manipulation intentionally alters occupancy (e.g., by removing occluders or revealing free space), confident semantic changes offer more direct indications of collateral object rearrangement.
Therefore, to preserve voxel-level semantic scene structure, we augment $\mathrm{OIG}$ with a Collateral Disturbance Constraint (CDC), which penalizes semantic changes in confident regions of the predicted post-action belief $\tilde{\Phi}_{t+1}^{a_t}$.
Using the semantic Dirichlet parameters $\lambda_t^S$, we represent the collateral disturbance as the set of disturbed confident voxels~$\mathcal{U}_\text{diff}$ as: 
\begin{equation}
\begin{aligned}
\chi^S_{t+1}(u)
&:=
\frac{N_{\mathrm{cls}}}
{\sum_{c=1}^{N_{\mathrm{cls}}} \eta_{t+1,u,c}^{S}}, \\
\mathcal{U}_\text{diff}(\Phi_t,\tilde{\Phi}_{t+1}^{a_t})
&:=
\left\{
u \mid
\chi^S_{t+1}(u) < \tau_\chi,\
\rho_{t+1,u} \neq \rho_{t,u}
\right\},
\end{aligned}
\label{eq:udiff}
\end{equation}
where $\tau_\chi$ is an uncertainty threshold and $\rho_{t,u}$ denotes the estimated semantic class at voxel $u$. 

Consequently, we constrain the number of disturbed confident voxels to remain below an allowable scene change threshold: $|\mathcal{U}_\text{diff}(\Phi_t, \tilde{\Phi}_{t+1}^{a_t})|<\Omega_c$. To incorporate this constraint into the action selection,  
we use a Lagrangian relaxation with multiplier $\zeta_\text{CDC}$, which yields the final objective:
\vspace{-2px}
\begin{equation}
\mathrm{DOIG}(a_t) :=
\mathrm{OIG}(a_t)
-
\zeta_\text{CDC}
|\mathcal{U}_\text{diff}(\Phi_t, \tilde{\Phi}_{t+1}^{a_t})|
\label{eq:doig}
\end{equation}
Notably, for pure viewpoint actions, $\mathrm{DOIG}$ reduces to $\mathrm{OIG}$, since no manipulation-induced disturbance is introduced. For grasp actions, the semantic region corresponding to the grasped object is excluded from the CDC, such that intentional removal is not penalized and only collateral changes to other confident regions are considered. Consequently, MS-MEM selects manipulation only when its expected occlusion reduction outweighs the predicted collateral disturbance.

\subsection{Uncertainty-Informed Grasp Selection}
\label{sec:UGS}
\begin{figure}[]
    \centering
    \includegraphics[width=\linewidth]{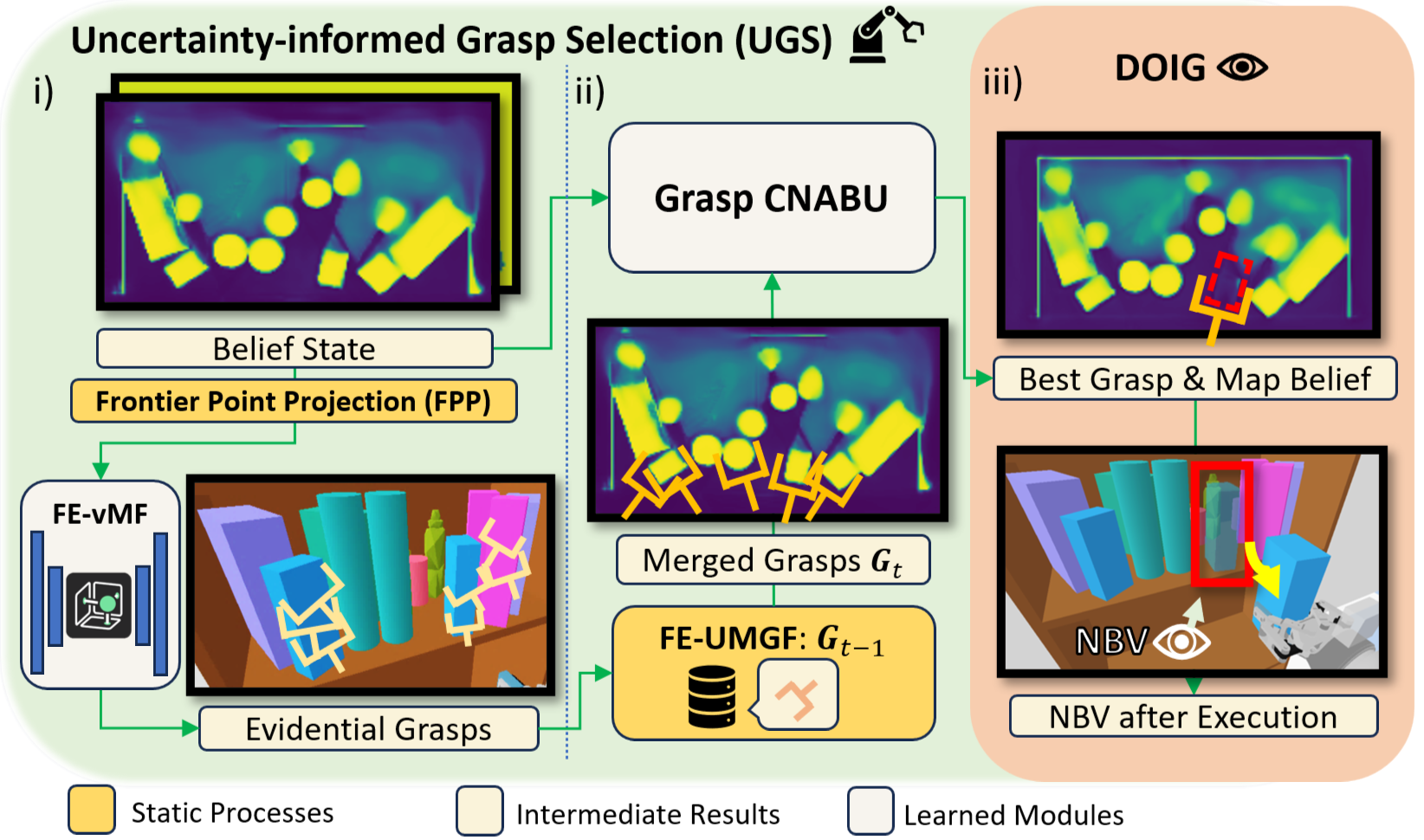}
    \caption{Uncertainty-Informed Grasp Selection (\textbf{UGS}) pipeline. 
    \textbf{(i)} Our proposed \textbf{FE-vMF} model predicts uncertainty-aware grasps from the current belief. 
    \textbf{(ii)} The grasp hypotheses are accumulated in a global grasp buffer using \textbf{FE-UMGF}.
    \textbf{(iii)} Our DOIG selects the grasp and subsequent NBV.}
    \vspace{-4mm}
    \label{fig:UGS}
\end{figure}
To extend the original MEM formulation~\cite{marques2025map} with selective occluder removal, we introduce the uncertainty-informed grasp selection pipeline, as illustrated in Fig.~\ref{fig:UGS}. 
Starting from the current occupancy belief~$\Phi_t^O$, our Full-Evidential vMF-Contact model (FE-vMF, Sec.~\ref{sec:FE-vMF}) predicts contact-level grasp hypotheses with evidential uncertainty. 
These hypotheses are then accumulated and refined over time using uncertainty-guided multi-view grasp fusion (Sec.~\ref{sec:FE-UMGF}).
Finally, for each executable grasp, a dedicated Grasp CNABU predicts the post-grasp belief $\tilde{\Phi}_{t+1}^{\boldsymbol{g}_t}$, enabling joint selection of the grasp and subsequent viewpoint that maximize the shared $\mathrm{DOIG}$ objective in Sec. \ref{sec:MSAS}.

\subsubsection{\textbf{Full-Evidential Grasp Learning}}
\label{sec:FE-vMF}

Our proposed FE-vMF extends vMF-Contact~\cite{shi2025vmf} from partial directional uncertainty to a full-evidential representation of the~$SE(3)$ grasp configuration. 
For each potential contact point $\boldsymbol{c}$, the model predicts uncertainty in both the grasp affordance and the grasp orientation, enabling downstream action selection to reason about grasp quality and confidence.

\begin{figure}[]
    \centering
    \includegraphics[width=\linewidth]{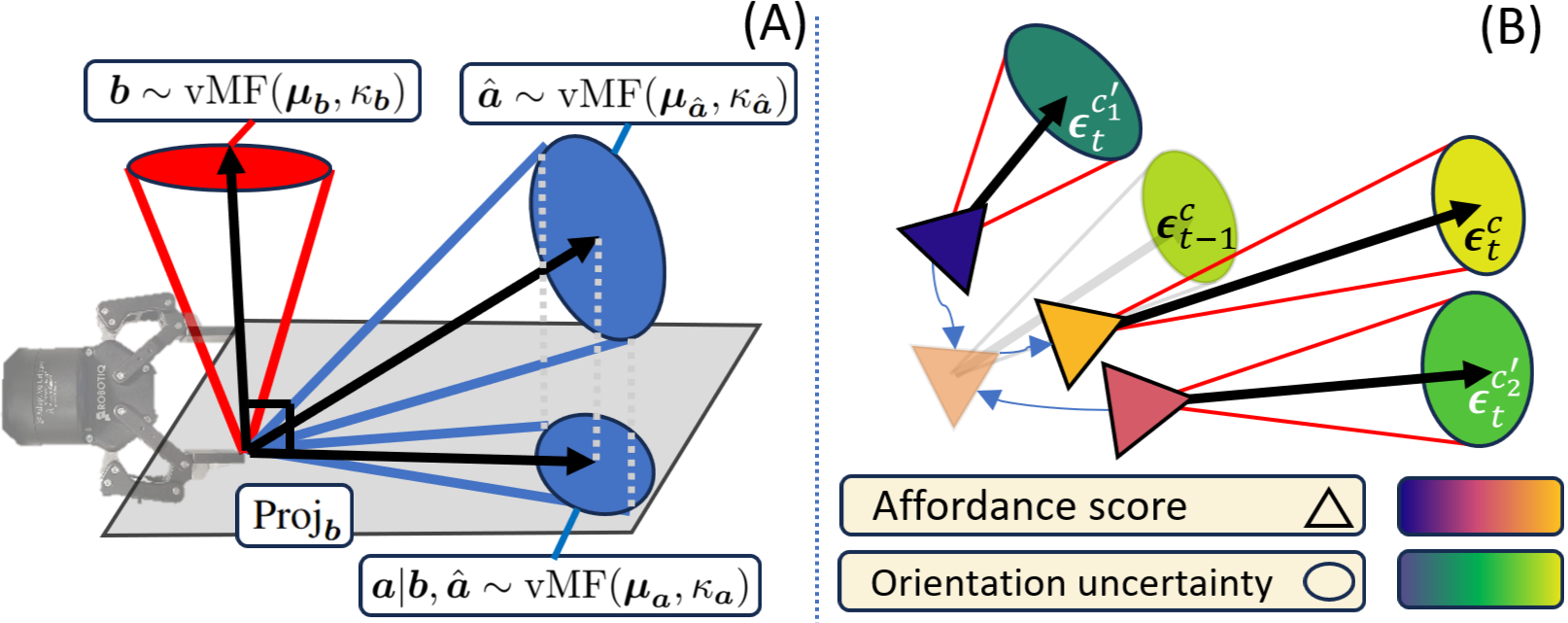}
        \caption{
    Illustration of the proposed evidential grasp representation and temporal fusion.
    \textbf{(A)} Uncertainty-aware orthogonal projection of the grasp orientation. The baseline direction $\boldsymbol{b}$ and the intermediate approach direction $\hat{\boldsymbol{a}}$ are modeled independently by von Mises--Fisher distributions. 
    \textbf{(B)} \textbf{FE-UMGF} pipeline. Current grasp predictions $\epsilon_t^{c'}$ are associated with a previous grasp hypothesis $\epsilon_{t-1}^{c}$. The matched hypotheses are fused into the updated representation~$\epsilon_t^{c}$, combining affordance evidence and orientation uncertainty across observations. Triangle color denotes grasp affordance, while ellipse color denotes orientation uncertainty.
}
    \label{fig:umgf}
    \vspace{-2mm}
\end{figure}

\paragraph{Dual Orientational Uncertainty Representation}
Unlike vMF-Contact~\cite{shi2025vmf}, which models uncertainty only for the baseline direction $p(\boldsymbol{b}| q,\boldsymbol{c})$, we represent both orientation components probabilistically. 
As shown in Fig.~\ref{fig:umgf}(A), the baseline $\boldsymbol{b}$ and an intermediate approach estimate $\hat{\boldsymbol{a}}$ are modeled as independent vMF distributions:
\begin{equation}
\hat{\boldsymbol{a}} \sim \mathrm{vMF}(\boldsymbol{\mu}_{\hat{\boldsymbol{a}}}, \kappa_{\hat{\boldsymbol{a}}}),
\quad
\boldsymbol{b} \sim \mathrm{vMF}(\boldsymbol{\mu}_{\boldsymbol{b}}, \kappa_{\boldsymbol{b}})
\end{equation}
The final approach distribution is obtained by projecting $\hat{\boldsymbol{a}}$ onto the subspace orthogonal to $\boldsymbol{b}$, thereby enforcing the grasp constraint $\boldsymbol{a}\perp\boldsymbol{b}$:
\begin{equation}
\boldsymbol{\mu}_{\boldsymbol{a}}
=
\frac{\mathrm{Proj}_{\boldsymbol{b}}\boldsymbol{\mu}_{\hat{\boldsymbol{a}}}}
{\left\|\mathrm{Proj}_{\boldsymbol{b}}\boldsymbol{\mu}_{\hat{\boldsymbol{a}}}\right\|},
\quad
\kappa_{\boldsymbol{a}}
=
\kappa_{\hat{\boldsymbol{a}}}
\left\|\mathrm{Proj}_{\boldsymbol{b}}\boldsymbol{\mu}_{\hat{\boldsymbol{a}}}\right\|,
\label{eq:proj}
\end{equation}
where $\mathrm{Proj}_{\boldsymbol{b}}=\boldsymbol{I}-\boldsymbol{\mu}_{\boldsymbol{b}}\boldsymbol{\mu}_{\boldsymbol{b}}^\top$ denotes the orthogonal projection matrix.
This formulation allows the model to express anisotropic orientation uncertainty. 
For instance, when two objects are closely positioned, contact points near their boundary may produce ambiguous grasp hypotheses that share a similar approach direction but have different baseline orientations, each corresponding to a grasp targeting a different object.

\paragraph{Evidential Affordance Representation}
Analogous to the evidential occupancy representation in Eq.~\eqref{eq:occ_beta}, we model the grasp affordance $q$ at contact point $\boldsymbol{c}$ as a Beta distribution,
$q^{\boldsymbol c}\sim\mathrm{Beta}(\alpha^{\boldsymbol c},\beta^{\boldsymbol c})$. 
The evidence parameters are predicted from raw logits $\hat{\boldsymbol e}^{\boldsymbol c}\in\mathbb{R}^2$ using a Softplus activation:
\begin{equation}
\alpha^{\boldsymbol{c}} = e^{\boldsymbol{c}}_\alpha + 1, 
\quad 
\beta^{\boldsymbol{c}} = e^{\boldsymbol{c}}_\beta + 1, 
\quad 
\boldsymbol e^{\boldsymbol{c}} = \mathrm{Softplus}(\hat{\boldsymbol{e}}^{\boldsymbol{c}}).
\label{eq:grasp_aff}
\end{equation}
The total evidence $S=\alpha^{\boldsymbol c}+\beta^{\boldsymbol c}$ determines both the predictive affordance and the associated epistemic uncertainty~\cite{gao2025comprehensive}:
\begin{equation}
    \mathbb{E}[q^{\boldsymbol c}] = \frac{\alpha^{\boldsymbol{c}}}{S}, 
    \quad 
    \chi^{\boldsymbol c} = \frac{2}{S}
\label{eq:occ_beta2}
\end{equation}
\paragraph{Learning \emph{FE-vMF}}
At each time step, we extract a front-surface point cloud from the expected occupancy belief~$\mathbb{E}[\Phi_t^O]$ using frontier point projection ($\mathrm{FPP}$):
\begin{equation}
\mathrm{pcd}=\mathrm{FPP}(\mathbb{E}[\Phi_t^O])
\in\mathbb{R}^{N_\mathrm{pcd}\times3}
\label{eq:frontier_projection}
\end{equation}
Here, ray casting isolates the currently visible surface points, and $N_{\mathrm{pcd}}=10^4$ points are sampled. 
The point cloud is processed by a small-scale Point Transformer~v3~(PTv3) backbone~\cite{wu2024point}, followed by an MLP that predicts the evidential grasp parameters for each contact point:
\begin{align}
\left\{\boldsymbol{f}^{\boldsymbol{c}_i}\right\}_{i=1}^{N_{\mathrm{pcd}}}
&=
\mathrm{PTv3}(\mathrm{pcd}),
\quad
\boldsymbol{\epsilon}^{\boldsymbol{c}_i}
=
\mathrm{MLP}(\boldsymbol{f}^{\boldsymbol{c}_i}), \\
\boldsymbol{\epsilon}^{\boldsymbol{c}_i}
&=
\{
\boldsymbol{\mu}^{\boldsymbol{c}_i}_{\hat{\boldsymbol{a}}},
\kappa^{\boldsymbol{c}_i}_{\hat{\boldsymbol{a}}},
\boldsymbol{\mu}^{\boldsymbol{c}_i}_{\boldsymbol{b}},
\kappa^{\boldsymbol{c}_i}_{\boldsymbol{b}},
\hat{\boldsymbol{e}}^{\boldsymbol{c}_i}
\}
\label{eq:FE-vMF_inference}
\end{align}
FE-vMF is trained with a composite evidential loss that supervises the orientation distributions for positive grasp contacts and the affordance evidence for all contacts:
%\begin{equation}
%\mathcal{L}^{\boldsymbol{c}_i}_\mathrm{FE}
%=
%q^{\dagger}
%\left(
%\mathcal{L}^\mathrm{BL}_{\boldsymbol{a}^{\dagger}}
%+
%\mathcal{L}^\mathrm{BL}_{\boldsymbol{b}^{\dagger}}
%\right)
%+
%\mathcal{L}^{\mathrm{EDL}}_{q^{\dagger}},
%\label{eq:fevmf_loss}
%\end{equation}
\begin{equation}\mathcal{L}^{\boldsymbol{c}_i}_\text{\emph{FE}} = q^{\dag}\Bigl(\mathcal{L}^\text{BL}_{\boldsymbol{a}^{\dag}} ({\boldsymbol{\mu}}^{\boldsymbol{c}_i}_{\hat{\boldsymbol{a}}}, \kappa^{\boldsymbol{c}_i}_{\hat{\boldsymbol{a}}}) + \mathcal{L}^\text{BL}_{\boldsymbol{b}^{\dag}}({\boldsymbol{\mu}}^{\boldsymbol{c}_i}_{{\boldsymbol{b}}},\kappa^{\boldsymbol{c}_i}_{b})\Bigr) + \mathcal{L}^{\text{EDL}}_{q^{\dag}} (\hat{\boldsymbol{e}}^{\boldsymbol{c}_i}),
 \label{eq:fevmf_loss}
 \end{equation} 
where $q^{\dagger}=\mathbb{I}(\boldsymbol c_i\approx\boldsymbol c^{\dagger})$ indicates whether $\boldsymbol c_i$ lies within 2 mm of a contact point $\boldsymbol c^{\dagger}$ associated with any ground-truth grasp.
The Bayesian orientation losses~$\mathcal{L}^\mathrm{BL}$ follow~\cite{shi2025vmf}, while the affordance term $\hat{\boldsymbol{e}}^{\boldsymbol{c}_i}$ uses the Evidential loss $\mathcal{L}^{\mathrm{EDL}}$ from~\cite{gao2025comprehensive}.

\subsubsection{\textbf{Full-Evidential Uncertainty-guided Multi-view Grasp Fusion}}
\label{sec:FE-UMGF}
Since grasp estimates are derived from partial observations, MEM’s active vision inherently allows the predicted grasp hypotheses to be progressively refined as observations from additional viewpoints become available. 
We therefore introduce Full-Evidential Uncertainty-guided Multi-view Grasp Fusion (FE-UMGF), a temporal fusion mechanism that integrates evidential grasp predictions across time and keeps them consistent with the evolving map belief.

Specifically, we maintain a global grasp buffer $\boldsymbol{G}_{t-1}$ and update it with the current predictions 
$\boldsymbol{G}'_t=\{\boldsymbol{\epsilon}^{\boldsymbol{c}'_i}\}_{i=1}^{N_{\mathrm{pcd}}}$:
\begin{equation}
    \boldsymbol{G}_t \leftarrow \emph{FE\text{-}UMGF}(\boldsymbol{G}_{t-1},\boldsymbol{G}'_t).
\end{equation}
As an extension of the grasp fusion from~\cite{shi2025viso}, we account for the full-evidential grasp representation. 
For a fused grasp at cluster center $\boldsymbol c$, the Beta evidence is updated by aggregating evidence from all newly assigned neighboring candidates $\boldsymbol c'\in C_t$, while decaying historical evidence with factor $\gamma$:
\begin{equation}
\alpha^{\boldsymbol{c}}_t
=
\gamma \alpha^{\boldsymbol{c}}_{t-1}
+
\sum_{\boldsymbol{c}'\in C_t}\alpha^{\boldsymbol{c}'}_{t},
\quad
\beta^{\boldsymbol{c}}_t
=
\gamma \beta^{\boldsymbol{c}}_{t-1}
+
\sum_{\boldsymbol{c}'\in C_t}\beta^{\boldsymbol{c}'}_{t}.
\end{equation}
The fused affordance is then computed using Eq.~\eqref{eq:occ_beta2}. 
The decay factor $\gamma$ gives higher relative weight to the current belief while preserving evidence accumulated from earlier views. 

For orientation, we follow~\cite{shi2025viso} while fusing both $\hat{\boldsymbol a}$ and $\boldsymbol b$ independently within each cluster. 
The final approach direction is recovered using the projection in Eq.~\eqref{eq:proj}.
\subsubsection{\textbf{Final Grasp Selection}}
After temporal fusion, each grasp hypothesis $\boldsymbol{\epsilon}^{\boldsymbol c}_t\in\boldsymbol G_t$ is converted into an executable grasp pose $\boldsymbol g_t^{\boldsymbol c}$ using the expectations of its evidential distributions. 
To predict the effect of executing this grasp, we use a grasp CNABU~$\sigma_g$, with the same architecture as the push CNABU~$\sigma_p$ in \cite{marques2025map}, to estimate the post-grasp evidential belief:
\begin{equation}
    \lambda_{t+1}^{\boldsymbol g_t^{\boldsymbol c}}
\leftarrow
\sigma_g(\lambda_t,\boldsymbol g_t^{\boldsymbol c}).
\end{equation}
The corresponding belief $\tilde{\Phi}_{t+1}^{\boldsymbol g_t^{\boldsymbol c}}$ is then used to evaluate the grasps and their best subsequent viewpoint under $\mathrm{DOIG}$, formulated in Eq.\eqref{eq:doig}:
\begin{equation}
(\boldsymbol{g}_t^*, v_{\boldsymbol{g}_t}^*)
=
\underset{
\boldsymbol{g}^{\boldsymbol{c}}_t \in\boldsymbol{G}_t,\,
v_{t+1} \in \mathcal{V}
}{\arg\max}
\ \mathrm{DOIG}(\boldsymbol{g}^{\boldsymbol c}_t).
\label{eq:grasp_ig}
\end{equation}

Similarly, for Uncertainty-informed Push Selection (UPS), a set of push candidates $\mathcal P_t$ is generated to reveal occluded regions following~\cite{dengler2025efficient}. 
For each $\boldsymbol p_t\in\mathcal P_t$, the push CNABU $\sigma_p$ predicts the post-push belief $\tilde{\Phi}_{t+1}^{\boldsymbol p_t}$ for $\mathrm{DOIG}$ evaluation.

\subsection{Multi-Skill Action Selection}
\label{sec:MSAS}

After the best candidates for grasping, pushing, and active view selection have been identified, our Multi-Skill Action Selection~(MSAS) strategy selects the best action to execute. 
Given the optimal grasp $\boldsymbol{g}_t^*$, push $\boldsymbol{p}_t^*$, and viewpoint $v_t^*$ candidate, each action is evaluated using its predicted post-action belief $\tilde{\Phi}_{t+1}^{a_t}$ and the shared $\mathrm{DOIG}$ objective~(Eq.~\eqref{eq:doig}) to determine the best overall action $a_t^*$:
\begin{equation}
    a_t^* =
    \underset{a_t \in \{\boldsymbol g^*_t, \boldsymbol p^*_t, v^*_t\}}
    {\arg\max}
    \ \mathrm{DOIG}(a_t)
\end{equation}
% This final comparison balances the expected reduction of map uncertainty against possible scene disturbance, enabling the system to choose between localized grasping, broader pushing, or additional sensing.
If a manipulation action is selected, the system subsequently executes the associated NBV~$v_{\boldsymbol{g}_t}^*$ or $v_{\boldsymbol{p}_t}^*$. 
The newly acquired observation updates the global evidential belief via observation CNABU $\sigma_o$ following Eq.~\eqref{eq:cnabu}. 

Similar to \cite{marques2025map}, once the scene is sufficiently mapped, as measured by the fraction of semantic map cells whose confidence exceeds $\tau_{\mathrm{conf}}$, manipulation actions are disabled, and the system proceeds with active view selection only. 
This process terminates when the step budget $T_{\max}$ is exhausted. %\todo{not entirely clear}

\label{sec:MSAS}

\section{Experiments}
We evaluate MS-MEM through a set of experiments designed to assess its mapping performance, multi-skill action selection, scene preservation, and uncertainty-aware grasp reasoning. 
Specifically, we investigate: (i) how pushing and grasping complement each other compared to single-skill baselines, (ii) how the CDC penalty affects the trade-off between information gain and scene disturbance, and (iii) how uncertainty is captured by the evidential grasp learning.

%\begin{figure}[]
%    \hspace{-3mm}
%    \includegraphics[width=1.03\linewidth]{pic/setup.png}
%    \vspace{-18px}
%    \caption{Simulation and real-world setups.}
%    \vspace{-14px}
%    \label{fig:setup}
%\end{figure}

\subsection{Training details}
\subsubsection{\textbf{Data Generation}}
The data generation pipeline is powered by the Pybullet simulation engine \cite{coumans2024python}, where the setup contains a confined shelf space with a UR5 robotic manipulator, equipped with a Robotiq 2F-85 gripper. %The simulation and real-world setups (Fig.~\ref{fig:setup}) are identical except for the perception system. 
In simulation, a set of projection-based pinhole cameras is used to approximate the wrist-mounted RealSense L515 camera. 
The predefined viewpoint candidates with a total number of $|\mathcal{V}|=300$ are kept consistent across both environments.

To train \emph{FE-vMF}, we generated $4\times10^3$ simulated shelf scenes. Each scene contained randomly arranged objects with an overall occupancy fraction between $30\%$ and $45\%$. We constructed the ground-truth grasp annotations in two stages. First, for each object model, we generated $10^5$ antipodal 6-DoF grasp candidates with the object placed in isolation. Second, after placing the objects in a shelf scene, we transformed the object-level grasp candidates into the scene coordinate frame and discarded any grasp that collided with the shelf or neighboring objects. The remaining collision-free grasps constitute the ground-truth grasp set $\{\boldsymbol g^\dagger\}$ for that scene. Each training sample consists of a point cloud obtained by merging observations from $3$--$10$ randomly selected viewpoints and is associated with approximately $100$--$400$ ground-truth grasps.

\subsubsection{\textbf{Training Details}} The training of the objective in Eq.~\eqref{eq:fevmf_loss} uses a single RTX 4090 GPU, optimized by AdamW \cite{loshchilov2017decoupled} with learning rate \mbox{$1e-5$}. 
The training of the grasp CNABU~$\sigma_g$ follows the same regime as the push CNABU $\sigma_p$ in~\cite{dengler2025efficient}: The dataset consists of $7000$ simulated grasp instances, maps before/after grasp execution, and volume-based action representation.
All CNABU networks are trained on a single \mbox{RTX A6000 GPU}. 
%The best validation performance was obtained after approximately $20$ epochs with a learning rate of $5 \times 10^{-4}$.

%\begin{figure}[]
%\centering
    %\includegraphics[width=.85\linewidth]{pic/manip-count.png}
    %\vspace{-5px}
    %\caption{Statistics on the number of each manipulation skill corresponds to the simulation experiments in Fig.\ref{fig:exp1}}
    %\vspace{-5px}
    %\label{fig:exp2}
%\end{figure}

\subsection{Simulation Experiments}
\label{sec:simulation}
We first evaluate MS-MEM in simulation to analyze its mapping performance and action selection behavior under controlled cluttered shelf scenarios. 
The evaluation was performed in the $25$ hand-crafted challenging scenes from~\cite{dengler2025efficient}.

\subsubsection{\textbf{Baselines and Metrics}}
We compare MS-MEM against three baselines to isolate the effect of multi-skill action selection and the disturbance-aware objective. 
\emph{Grasp Only} uses grasping as the only manipulation primitive, while \mbox{\emph{Push Only}} corresponds to the original MEM framework~\cite{marques2025map}, using pushing and active viewpoint selection. 
To show the benefit of our proposed $\mathrm{DOIG}$ metric, we also compare our method against \emph{w/o CDC}, a variant of MS-MEM that selects the actions via the standard $\mathrm{OIG}$ objective without CDC penalty. 
Finally, \emph{Ours} denotes the full MS-MEM system. Active view planning remains enabled for all baselines.

We evaluate mapping performance using occupancy and semantic Intersection over Union (IoU) between final map outcomes and ground truths. 
To quantify scene preservation, we measure the cumulative displacement of all objects from their initial positions (Position Change). For grasped objects, we measure only the displacement of any object except the grasped one, such that the metric measures scene disturbance rather than the intended removal. Moreover, the number of executed actions for each step is recorded to compare between multi-skill synergy and single-skill baselines.

\subsubsection{\textbf{Mapping Performance}}
\begin{figure}[]
    \includegraphics[width=.98\linewidth]{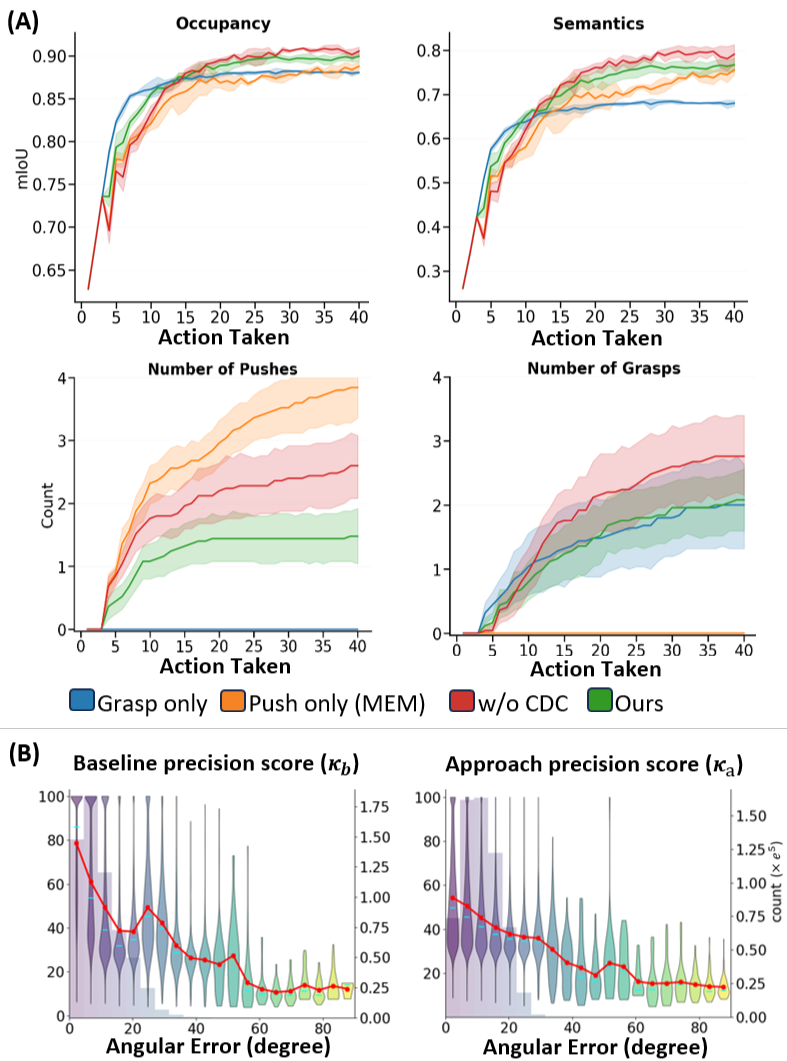}
        \caption{
    Quantitative simulation and uncertainty-calibration results.
    \textbf{(A)} Mapping performance and manipulation statistics for the baselines described in Sec.~\ref{sec:simulation}. 
    %Multi-skill variants achieve higher mapping accuracy than the single-skill baselines, while the CDC reduces the use of disruptive push actions.
    \textbf{(B)} Calibration of the predicted directional precision for the grasp baseline direction $\boldsymbol{b}$ and approach direction~$\boldsymbol{a}$.}
    \label{fig:exp1}
    %\caption{(A) Simulation experiments under baselines in Sec.~\ref{sec:simulation}; (B) Statistics on the uncertainty calibration of directional precision ($\kappa_{\boldsymbol{b}}$, $\kappa_{\boldsymbol{a}}$) %wrt. ground truths directional errors in the test dataset.*Note: The right axis shows the sample count in units of $1e5$.
    %}
    \vspace{-2mm}
\end{figure}

Fig.~\ref{fig:exp1} (A) and Table~\ref{tab:metrics_last_step_transposed} summarize the mapping performance of all methods. Overall, in terms of mapping accuracy, methods that combine pushing and grasping outperform the single-skill baselines, with both \emph{Ours} and \emph{w/o CDC} achieving the highest occupancy IoU and clearly improved semantic IoU. This demonstrates the complementary benefits of pushing to expose hidden regions, and grasping to remove selected occluders once they become reachable.

This interpretation is further supported by the detailed manipulation action statistics in Fig.~\ref{fig:exp1} (A). 
Here, the \mbox{\emph{w/o CDC}} variant performs more grasps than even the \emph{Grasp Only} baseline, suggesting that pushing actively creates grasp opportunities by separating clutter, revealing object boundaries, and increasing free space. 
Thus, the benefit of multi-skill reasoning is not only the availability of more actions, but also from sequencing complementary skills so that each improves the effectiveness of the other.

Finally, the comparison between \emph{w/o CDC} and \emph{Ours} highlights the effect of the proposed disturbance-aware objective. 
While both methods benefit from push-grasp synergy, adding CDC substantially reduces object displacement while maintaining strong mapping performance. 
In particular, CDC reduces the average number of pushes more strongly than grasps, suggesting that it primarily suppresses disruptive rearrangements while retaining grasping as a more localized manipulation primitive. 
As a result, the full MS-MEM achieves a better balance between semantic mapping accuracy and scene preservation than \emph{w/o CDC}.

\begin{table}[t]
\centering
\caption{Mapping performance at the last step ($t=40$).}
\label{tab:metrics_last_step_transposed}
\begin{tabular}{lccc}
\toprule
\makecell[c]{Method}
& \makecell[c]{Occupancy\\mIoU$\uparrow$}
& \makecell[c]{Semantics\\mIoU$\uparrow$}
& \makecell[c]{Position\\Change (m)$\downarrow$} \\
\midrule
Grasp only
& $0.880 \pm 0.001$
& $0.681 \pm 0.006$
& $0.228 \pm 0.018$ \\

Push only~\cite{marques2025map}
& $0.887 \pm 0.005$
& $0.756 \pm 0.011$
& $1.571 \pm 0.094$ \\

w/o CDC
& $0.905 \pm 0.005$
& $0.791 \pm 0.022$
& $1.231 \pm 0.189$ \\

Ours
& $0.899 \pm 0.006$
& $0.767 \pm 0.016$
& $0.707 \pm 0.094$ \\
\bottomrule
\end{tabular}
\vspace{-3mm}
\end{table}

% Requires \usepackage{multirow}

% \begin{table}[t]
% \centering
% \caption{Uncertainty calibration performance on orientation. \todo{are the results significant?}}
% \label{tab:metrics_side_by_side}
% \setlength{\tabcolsep}{4.7pt}
% \begin{tabular}{llccccc}
% \toprule
% Method
% & Orientation
% & NLL$\downarrow$
% & \makecell[c]{AUSC\\AL$\downarrow$}
% & \makecell[c]{AUSE\\AL$\downarrow$}
% & \makecell[c]{AUSC\\EP$\downarrow$}
% & \makecell[c]{AUSE\\EP$\downarrow$} \\
% \midrule
% \multirow{2}{*}{vMF-Contact}
% & Baseline
% & $-0.69$
% & $11.09$
% & $3.92$
% & $10.10$
% & $4.06$ \\
% & Approach
% & /
% & /
% & /
% & /
% & / \\
% \midrule
% \multirow{2}{*}{FE-vMF}
% & Baseline
% & $-0.96$
% & $9.46$
% & $3.02$
% & $10.94$
% & $4.45$ \\
% & Approach
% & $-1.15$
% & $7.13$
% & $2.42$
% & $9.61$
% & $4.88$ \\
% \bottomrule
% \end{tabular}
% \end{table}

\begin{table}[t]
\centering
\caption {Real-world object recognition performance.}
\label{tab:real_world_results}
\setlength{\tabcolsep}{1pt}
\begin{tabular}{lccccc}
\toprule
\makecell[c]{Method} 
& \makecell[c]{Correctly\\Found$\uparrow$} 
& \makecell[c]{Misclassified\\But Found}$\downarrow$ 
& \makecell[c]{Not\\Found}$\downarrow$
& \makecell[c]{Hallucinated$\downarrow$} 
& \makecell[c]{Position \\ Change (m)$\downarrow$} \\
\midrule
Grasp only & $38$ & $19$ & $16$ & $18$ & $0.198 \pm 0.017$ \\
Push only~\cite{marques2025map} & $40$ & $23$ & $10$ & $16$ &$ 0.567 \pm 0.045$ \\
Ours & $44$ & $21$ & $8$ & $15$ & $0.408 \pm 0.037$\\
\bottomrule
\end{tabular}
\vspace{-3mm}
\end{table}

\subsubsection{\textbf{Evidential Grasp Learning Performance}}

\begin{figure*}[]
    \centering
    \includegraphics[width=\linewidth]{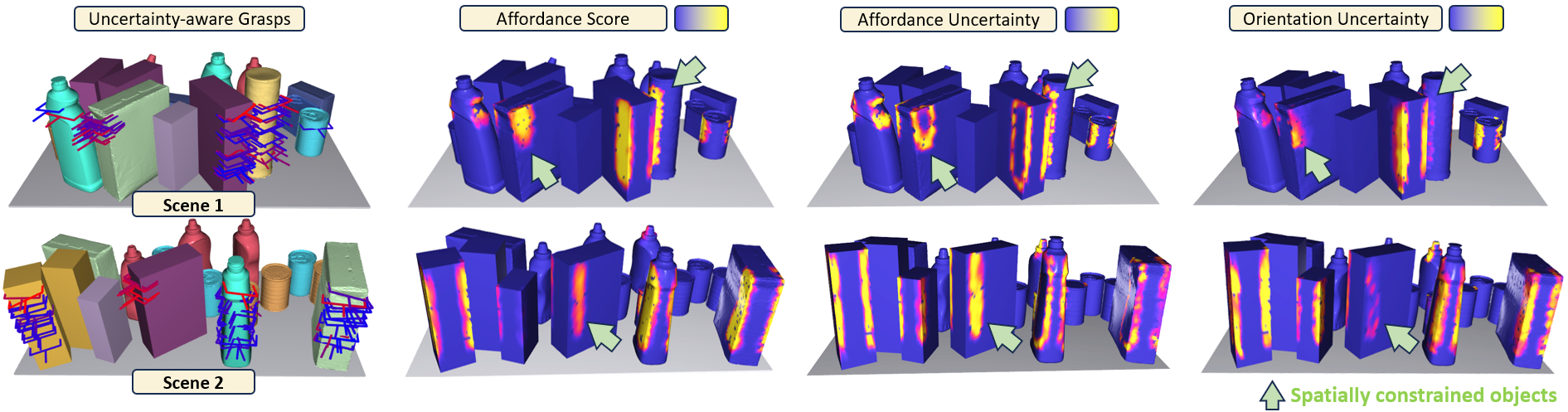}
    \vspace{-6mm}
    \caption{Qualitative visualization of the proposed FE-vMF predictions in two cluttered shelf scenes. From left to right, the columns show the predicted uncertainty-aware grasps, grasp affordance score $q^{\boldsymbol{c}}$, affordance uncertainty $\chi^{\boldsymbol{c}}$ from Eq.~\eqref{eq:occ_beta2}, and orientation uncertainty visualized as $\bigl(\kappa_{\hat{\boldsymbol{a}}}^{\boldsymbol{c}_i}+\kappa_{\boldsymbol{b}}^{\boldsymbol{c}_i}\bigr)^{-1}$. Brighter colors indicate larger values. %The highlighted examples illustrate that partially observed or collision-sensitive regions tend to exhibit high affordance uncertainty, whereas objects whose feasible grasp directions are strongly restricted by neighboring geometry exhibit lower orientation uncertainty.
    }
    \label{fig:unc_grasp}
    \vspace{-3mm}
\end{figure*}

\begin{figure}[]
    \includegraphics[width=\linewidth]{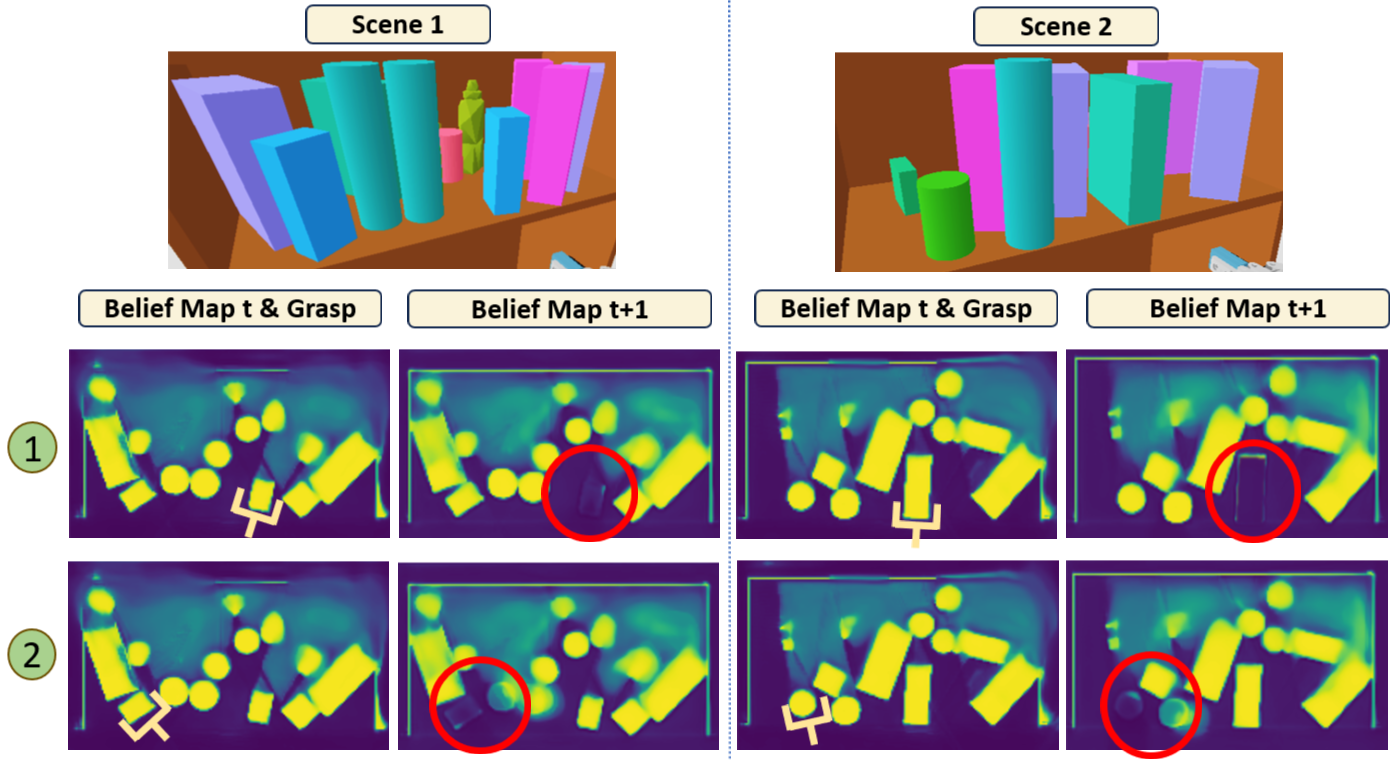}
    
    \caption{Qualitative predictions of the Grasp CNABU in two cluttered shelf scenes. For each scene, the left column shows the current belief map and a candidate grasp, while the right column shows the corresponding predicted post-grasp belief at time $t+1$. Candidate~\cnum{1} removes the selected object with limited influence on its surroundings, resulting in a localized and comparatively confident belief update. In contrast, candidate~\cnum{2} is likely to contact or displace neighboring objects during execution. 
    %The Grasp CNABU captures this possible collateral disturbance as increased uncertainty and spatially broader belief changes in the highlighted regions.
    }
    \label{fig:UGS_out}
\end{figure}

Fig. \ref{fig:unc_grasp} shows the visualization of FE-vMF outcomes.
Here, high affordance uncertainty primarily occurs at partially observed contact points, in collision-sensitive regions during grasping, or where the grasp affordance is intrinsically ambiguous ($q^\mathbf{c}\approx0.5$). Orientation uncertainty, visualized as $\bigl(\kappa_{\hat{\boldsymbol{a}}}^{\boldsymbol{c}_i}+\kappa_{\boldsymbol{b}}^{\boldsymbol{c}_i}\bigr)^{-1}$, increases when multiple collision-free approach directions are plausible for less physically constrained objects. Conversely, when nearby obstacles constrain the feasible approach (green arrows in Fig.~\ref{fig:unc_grasp}), directional uncertainty is reduced by restricting the grasp orientations to avoid collision. The quantitative uncertainty calibration statistics is depicted in Fig.~\ref{fig:exp1} (B), where lower predicted precision for larger aregular errors demonstrates that the orientation estimates accurately capture directional uncertainty.

\subsection{Real-World Performance}

Finally, we evaluate whether the proposed MS-MEM action-selection pipeline transfers zero-shot from simulation to a physical shelf setup, operated by a UR5 manipulator equipped with a Robotiq 2F-85 end effector and a Realsense L515 camera. For quantitative evaluation, we additionally use the Vicon motion capture system to track all object poses. The object poses before/after runtime are then used to generate ground-truth semantic occupancy maps for final evaluation, including object recognition metrics (following \cite{marques2025map}) and physical position change. In total, the real-world evaluation includes $5$~challenging scenes, each containing $69$ objects.

Table~\ref{tab:real_world_results} shows that \emph{Ours} achieves the best performance on correct object recognition among the baselines, with $44$~indentified objects in total compared with $40$ for \emph{Push only} and $38$~for \emph{Grasp only}. The improvement is mainly contributed by reduced missed objects, where \emph{Ours} leaves only $8$ objects not found, whereas other baselines miss over $10$. The displacement results further support the intended trade-off. \emph{Grasp only} causes the least object disturbance (i.e., Position Change), but its conservative interaction leaves many occluded objects unresolved. In contrast, \mbox{\emph{Push only}} improves coverage, but produces the largest scene disturbance, with an average displacement of $0.567$ m. \emph{Ours} achieves the best object recognition performance while reducing Position Change by $28.0\%$ compared with \emph{Push only}, from $0.567\,\mathrm{m}$ to $0.408\,\mathrm{m}$. Moreover, the grasp success rate of \emph{Ours} is improved by $30\%$ compared to \emph{Grasp only}. %This indicates that MS-MEM does not rely on disruptive pushing alone; instead, it selectively combines observation, pushing, and grasping to reveal uncertain regions while preserving already confident parts of the scene belief.
These results show that combining pushing and grasping improves object recognition and grasp success while reducing scene disturbance compared with the single-skill baselines.

\section{Conclusion}
In this paper, we presented MS-MEM, an evidential framework for uncertainty-aware mapping of confined cluttered spaces via active decision-making under grasp, push, and active view selection. By extending MEM~\cite{marques2025map} with the proposed FE-vMF for evidential grasp representation learning, FE-UMGF grasp fusion, and the unified $\mathrm{DOIG}$ objective, \mbox{MS-MEM} enables direct comparison of heterogeneous action skills under a shared evidential belief representation. This supports action selection and balances occlusion reduction against collateral scene disturbance. Our results show that jointly leveraging pushing and grasping provides clear advantages over single-skill baselines for evidential mapping.

\section{Acknowledgement}

GPT-5.5 was used for text enhancement and proofreading.

\bibliographystyle{IEEEtran}
\bibliography{ref}
\end{document}